\documentclass[10pt, journal]{IEEEtran}  

\IEEEoverridecommandlockouts     

\usepackage{color,amsmath,amstext,amsfonts,amssymb, mathrsfs,mathtools}
\usepackage{graphicx,algorithm,algorithmic}
\usepackage[algo2e,linesnumbered]{algorithm2e}
\usepackage{tikz}
\usepackage{setspace}
\usepackage{array} 
\usepackage{booktabs}  
\usepackage{bbm}
\usepackage{wasysym}
\usepackage{textcomp}
\usepackage{hyperref} 
\usepackage[sort,compress]{cite}

\newtheorem{lemma}{Lemma}
\newtheorem{assumption}{Assumption}

\newtheorem{theorem}{Theorem}

\newtheorem{remark}{Remark}

\renewcommand{\thesubsubsection}{{\it \Alph{subsection}.\arabic{subsubsection}.}}

\SetKwInput{KwInput}{Input}

\newcounter{l1}
\newcounter{l2}
\newcounter{l3}
\newcommand{\bdotlist}{\begin{list}{$\bullet$}{}}
\newcommand{\bboxlist}{\begin{list}{$\Box$}{}}
\newcommand{\bbboxlist}{\begin{list}{\raisebox{.005in}{{\tiny $\blacksquare$ \ \ }}}{}}
\newcommand{\bdashlist}{\begin{list}{$-$}{} }
\newcommand{\blist}{\begin{list}{}{} }
\newcommand{\barablist}{\begin{list}{\arabic{l1}}{\usecounter{l1}}}
\newcommand{\balphlist}{\begin{list}{(\alph{l2})}{\usecounter{l2}}}
\newcommand{\bAlphlist}{\begin{list}{\Alph{l2}.}{\usecounter{l2}}}
\newcommand{\bdiamlist}{\begin{list}{$\diamond$}{}}
\newcommand{\bromalist}{\begin{list}{(\roman{l3})}{\usecounter{l3}}}

\providecommand{\norm}[1]{\lVert#1\rVert}

\newcommand{\beq}{\begin{equation}}
\newcommand{\eeq}{\end{equation}}

\let\[\equation
\let\]\endequation

\newcommand{\tn}{\textnormal}

\DeclarePairedDelimiter\floor{\lfloor}{\rfloor}
\DeclarePairedDelimiterX{\Norm}[1]{\lVert}{\rVert}{#1}

\DeclareMathOperator*{\argmin}{arg\,min}

\allowdisplaybreaks

\title{ Meta-Learning Online Control for \\Linear Dynamical Systems}

\author{Deepan Muthirayan, Dileep Kalathil, and Pramod P. Khargonekar
\thanks{This work is supported in part by the National Science Foundation under Grant ECCS-1839429 and NSF- CAREER-EPCN-2045783. 
D. Muthirayan and P. P. Khargonekar are with the Department of Electrical Engineering and Computer Sciences, University of California Irvine, Irvine, CA (emails: deepan.m@uci.edu, pramod.khargonekar@uci.edu). Dileep Kalathil is with the Department of Electrical and Computer Engineering, Texas A\&M University (email:dileep.kalathil@tamu.edu).}
}

\begin{document}

\maketitle

\begin{abstract}

In this paper, we consider the problem of finding a meta-learning online control  algorithm that can learn across the tasks when faced with  a sequence of $N$ (similar) control tasks.  Each  task involves controlling a linear dynamical system for a finite horizon of $T$ time steps. The cost function and system noise at each time step are adversarial and unknown to the controller before taking the control action. Meta-learning is a broad approach where the goal is to prescribe an online policy for any new unseen task exploiting the information from other tasks and the similarity between the tasks. We propose a meta-learning online control algorithm for the control setting and characterize its performance by \textit{meta-regret}, the average cumulative regret across the tasks. We show that when the number of tasks are sufficiently large, our proposed approach achieves a meta-regret that is smaller by a factor $D/D^{*}$ compared to an independent-learning online control algorithm which does not perform learning across the tasks, where $D$ is a problem constant and $D^{*}$ is a scalar that decreases with increase in the similarity between tasks. Thus, when the sequence of tasks are similar the regret of the proposed meta-learning online control is significantly lower than that of the naive approaches without meta-learning. We also present experiment results to demonstrate the superior performance  achieved by our meta-learning algorithm.   

\end{abstract}

\section{Introduction}
\label{sec:int}

Meta-learning is a powerful paradigm in machine learning for  \textit{learning-to-learn} new tasks efficiently, e. g., with limited data \cite{thrun2012learning}. Meta-learning is based on the  intuitive idea that if the new task is similar to previous tasks, it can be learned very quickly by using the data and knowledge from previously encountered related tasks. Recently there has been tremendous progress in practical algorithms for meta-learning \cite{finn2017model, ravi2017optimization, nichol2018first} with impressive performance in many applications such as image classification \cite{wang2020frustratingly}, natural language processing \cite{bragg2021flex}, and robotic control  \cite{NagabandiCLFALF19}. These algorithms, however, are in the \textit{batch learning} setting, where data sets composed of different tasks are available for offline training. A meta-model (typically a neural network) is then trained using these data sets with the objective of fast adaptation to a new/unseen  task at the test time using only a few  data samples corresponding to that new task. Significantly different from the batch learning setting which are {offline} by nature, many learning algorithms  have to operate in an \textit{online } setting where the data samples are obtained in a sequential manner. 
For example, personalized recommendation systems \cite{li2010contextual}, various applications in robotics \cite{lesort2020continual, petrivc2014online, alambeigi2018robust, romeres2019derivative, tesfazgi2021personalized}, demand response management in smart grid \cite{kalathil2015online}, and load balancing in data centers \cite{lin2012online} require online learning. 

Online convex optimization (OCO)  \cite{shalev2011online, hazan2019introduction} focuses on developing algorithms for  online learning setting where the loss functions are sequentially revealed and the learner is trained as well as tested at each round. The standard OCO objective is to minimize the  regret which is defined as the difference between the cumulative cost incurred by the online algorithm and  the optimal policy from a certain class of policies. Even though the OCO approach offers a fundamental theoretical framework to analyze a variety of online learning scenarios, the existing works do not consider how the past experience can be used to accelerate adaptation to a new task, which is the key idea behind meta-learning. There are many works in the area of  online control algorithms for dynamical systems  with uncertain/unknown disturbances, system parameters and cost functions. The online control literature extends the OCO approach to problems with dynamics 
\cite{dean2018regret,  mania2019certainty, agarwal2019online, simchowitz2020improper}. However, these existing works only consider the problem of learning \textit{within} a task assuming that the task is fixed. In particular, they do not consider the possibility of \textit{learning across the tasks} when faced with a sequence of similar control tasks.

In this paper, we consider the problem of finding a meta-learning online control  algorithm that learns across the tasks when faced with  a sequence of $N$ (similar) control tasks.  Each  task involves controlling a linear dynamical system for $T$ time steps. The cost function and system noise at each time step are adversarial and unknown to the algorithm before taking the control action. The primary role of a meta-learning algorithm is to prescribe an online control policy for any new unseen task  exploiting the information from prior tasks and the similarity between the tasks. We characterize the performance of a meta-learning online control algorithm by \textit{meta-regret}, the average (taken over the tasks) cumulative regret across the tasks. Our goal is to develop a meta-learning online control algorithm that can achieve superior performance, in theory and practice,  over an independent-learning online control algorithm which applies a standard online control algorithm  to each task without performing any learning across the tasks.   

Our approach is motivated by some  recent works in online meta-learning \cite{finn2019online, balcan2019provable, khodak2019adaptive} which combine the meta-learning idea with the OCO framework.  In \cite{finn2019online}, the authors extend the model-agnostic meta-learning (MAML) approach \cite{finn2017model} to the online setting. Their goal is to learn a good meta-policy parameter that allows fast adaptation to all the previously seen tasks by taking only a few gradient steps from this meta-policy parameter. The work that is closest ours is  \cite{balcan2019provable}, which proposes the  Follow-the-Meta-Regularized-Leader (FTMRL) approach. FTMRL learns a meta-intialization for a task specific OCO algorithm 
such that the individual task regret of these algorithms improves with the similarity of the online tasks. However, these works consider only the online optimization setting without state evolution. In particular, they do not consider the more challenging problem of online control for uncertain dynamical systems. 

\textbf{Our contributions:} We consider the problem of developing a meta-learning online control algorithm for a sequence of similar control tasks. Each  task involves controlling a linear dynamical system with adversarial cost functions and disturbances, which are unknown before taking the control action. Our algorithm has a two loop structure where the outer loop performs the meta-learning update to prescribe an initialization parameter for the task specific online control algorithm used in the inner loop. We show that when the number of tasks are sufficiently large the meta-regret of our proposed approach is smaller by a factor $D/D^{*}$ compared to an independent-learning online control algorithm which does not perform learning across the tasks, where $D$ is a problem constant and $D^{*}$ is a scalar that represents the task similarity ($D^{*}$ decreases with similarity between tasks). Therefore, when the sequence of tasks are similar, i.e., when $D^{*} \ll D$, we achieve a regret that is significantly lower than that of the naive approaches without meta-learning. We also present experiments results to demonstrate the superior performance of our meta-learning algorithm. 

Our technical contribution lies in expanding the framework and technical analysis of online control to incorporate meta-learning. To the best of our knowledge, ours is the first work that combines the ideas of meta-learning and online control to develop a learning algorithm with provable guarantees for its performance. The conference version of this paper presents a simpler algorithm that assumes the knowledge of $D^{*}$. In this version, we introduce a general algorithm that does not require the knowledge of $D^{*}$.


\vspace{0.3cm}

\noindent \textbf{Related Works:}

\textit{Online Control:} Substantial number of works have been published in the area of online control \cite{dean2018regret,  mania2019certainty, agarwal2019online, simchowitz2020improper, cohen2019learning, simchowitz2020naive, hazan2020nonstochastic}. Most of these works focus on developing online control algorithms for linear dynamical systems with provable guarantees for the regret. In our work we make use of the task specific online control algorithm proposed in \cite{agarwal2019online}. This considers the control of a known linear dynamic system with adversarial disturbance and (convex) cost functions and shows that the proposed algorithm can achieve $\mathcal{O}(\sqrt{T})$ regret for a given task. Our meta-learning online control algorithm is developed by extending the task specific  online control algorithm proposed in \cite{agarwal2019online} with an additional outer loop for performing the meta-learning update and slightly modifying the  task specific (inner loop) update.

\textit{Adaptive and Robust Control:} Classical adaptive and robust control literature addresses the problem of control of systems with parametric, structural, modeling and disturbance uncertainties \cite{sastry2011adaptive, aastrom2013adaptive, ioannou2012robust, ZhouDoyleGlover-RobustControl-Book}. Typically, these classical approaches are concerned with stability and asymptotic performance guarantees of the systems. Online control literature focuses typically on the finite time regret performance of the algorithms. This is one of the key differences compared to the conventional adaptive and robust control literature, and it requires combining techniques from statistical learning, online optimization and control. In this work, we focus on the online control approach for developing our meta-learning algorithm.

\textbf{Notations:} Unless otherwise specified $\norm{\cdot}$ denotes the Euclidean norm and the Frobenious norm for vectors and matrices respectively. We use $\mathcal{O}(\cdot)$ for the standard big-O notation while $\widetilde{\mathcal{O}}(\cdot)$ denotes the big-O notation neglecting the poly-log terms. We also use $o(\cdot)$ for the standard little-o notation. Further, when a function $g(n) = o_n(1)$, then $g(n) \rightarrow 0$ as $n \rightarrow \infty$. We denote the sequence $(x_{m_1}, x_{m_1+1}, \dots , x_{m_2})$ compactly by $x_{m_{1}:m_{2}}$. 

\section{Problem Setting}
\label{sec:form}

We consider the problem of finding a meta-learning online control (M-OC) algorithm that learns across the tasks when faced with  a sequence of (similar) control tasks. The sequence of tasks are denoted as $\tau_1, \tau_2, \dots, \tau_N$. Each control task $\tau_{i}$ involves controlling a linear dynamical system for $T$ time steps whose system dynamics is given by the equation 
\begin{align}
 x_{i,t+1} = A_i x_{i,t} + B_i u_{i,t} + w_{i,t}, ~1 \leq t \leq T,
\label{eq:syseq}   
\end{align}
where $A_i \in \mathbb{R}^{n\times n}$ and $B_i \in \mathbb{R}^{n \times m}$ are the matrices that paramaterize the system, and  $x_{i,t} \in \mathbb{R}^n$ is the state, $u_{i,t} \in \mathbb{R}^m$ is the action,  $w_{i,t} \in \mathbb{R}^n$ is the system noise at time $t$. For conciseness we represent the system parameter for task $\tau_{i}$ as $\theta_i = [A_i, B_i]$. We assume that the systems noise is adversarial.

A control policy $\pi$ for task $\tau_i$ selects a control action $u^{\pi}_{i,t}$ at each time $t$ depending on the available information, resulting in a sequence of actions $u^{\pi}_{i,1:T}$ and the state trajectory $x^{\pi}_{i,1:T}$. The cumulative cost of a policy $\pi$ under the system dynamics \eqref{eq:syseq} is given by 
\begin{align}
\label{eq:cumulative-cost}
 {J}_{i}(\pi) = \sum_{t = 1}^T c_{i,t}(x^{\pi}_{i,t},u^{\pi}_{i,t}),
\end{align}
where $c_{i,t}$ is the cost function for task $\tau_{i}$ at time $t$. We assume that $c_{i,t}$s are arbitrary convex functions. The typical goal is to find the optimal policy $\pi^{\star}_{i}$ such that $\pi^{\star}_i = \argmin_{\pi} J_{i}(\pi)$. Clearly, computing $\pi^{\star}_{i}$ requires the knowledge of the system parameter $\theta_i$ and the entire sequence of cost functions $c_{i,1:T}$. 

The  online control framework considers the more realistic setting where  the future cost functions are not available for deciding the control action $u_{i,t}$ at time $t$. More precisely the policy $\pi_i$ for task $\tau_{i}$ has only the following information at each time $t$ for selecting the action $u_{i,t}$: $(i)$ past and current state observations $x_{i,1:t}$, $(ii)$ past control actions $u_{i,1:t-1}$, $(iii)$ past cost functions $c_{i,1:t-1}$. We also assume that the system parameter $\theta_{i}$ is known to the control policy. 
The \textbf{task regret} of the control policy $\pi_i$ for the task $\tau_{i}$ is defined as
\beq 
R^i_T (\pi_i) =  J_i(\pi_i) - \min_{\pi \in \Pi} J_i(\pi),
\label{eq:regret}
\eeq 
where $\Pi$ is the class of control policies. The objective is to find a policy that minimizes the task regret assuming that the task is fixed. In particular the existing online control  algorithms do not consider \textit{learning across tasks} when faced with a sequences of similar control tasks. 

\textit{Our goal is to find a \textit{meta-policy} $\pi^{\mathrm{m}}$ that can learn across the tasks when faced with a sequence  of (similar) control tasks $\tau_1, \tau_2, \dots, \tau_N$ and minimize the task regret for individual tasks.} A meta-policy $\pi^{\mathrm{m}}$ produces a sequence of task specific policies $\pi^{\mathrm{m}}_{i}, 1 \leq i \leq N,$  by learning across the tasks. For deciding the task specific policy $\pi^{\mathrm{m}}_{i}$ for task $\tau_{i}$ the meta-policy $\pi^{\mathrm{m}}$ makes use of the observation available from the previous tasks: the state observations, cost functions, and task specific policies for all previous tasks $j \leq i-1$. Since the objective of the meta-policy is to generate task specific policies which can do well on individual tasks, the performance of the meta-policy is characterized by the metric \textbf{meta-regret}, formally defined as
\beq 
{R}^{\mathrm{meta}}_N(\pi^\mathrm{m}) = \frac{1}{N}\sum_{i = 1}^N R^i_T(\pi^{\mathrm{m}}_{i}).
\label{eq:averageregret}
\eeq 
Our objective is to find a meta-policy that performs better than an \textit{independent-learning  online control algorithm} which applies a standard online control algorithm independently to each task without performing any learning across the tasks. 

We make the following assumptions. Please note that the assumptions stated below are standard in the (task specific) online control literature \cite{agarwal2019online} and no further assumptions are made. 
\begin{assumption}[System Model]
(i) The system matrices for each task  are bounded,  $\norm{A_i} \leq \kappa_A$, and $\norm{B_i} \leq \kappa_B$, where $\kappa_A$ and $\kappa_B$ are constants. (ii) The disturbance at time $t$ of any task  is  bounded, $\norm{w_{i,t}} \leq \kappa_{w}$, where $ \kappa_{w}$ is a constant.
\label{ass:sys}
\end{assumption}
\begin{assumption}[Cost Functions]
\label{ass:cost}
For all tasks $i, 1 \leq i \leq N$  and all time steps $t, 1 \leq t \leq T$,
(i) the costs functions $c_{i,t}$s are convex, (ii) for any $x$ and $u$ with $\norm{x}\leq S, \norm{u} \leq S$, 
\beq 
\norm{c_{i,t}(x,u)} \leq \beta S^2, \norm{\nabla_x c_{i,t}(x,u)}, \norm{\nabla_u c_{i,t}(x,u)} \leq GS, \nonumber 
\eeq 
\end{assumption}

\section{Review: Online Control Algorithm}
\label{sec:singletaskcont}

In this section we give a brief description of the task specific online control (OC) algorithm proposed in \cite{agarwal2019online}. We drop the task subscript $i$ because the discussion here is for a single task. Our meta-learning online control algorithm is developed by extending the task specific OC algorithm with an additional outer loop for performing the meta-learning update and appropriately modifying the  task specific (inner loop) update.  

The OC algorithm proposed in  \cite{agarwal2019online} uses a  control policy parameterized by two matrices, a fixed matrix $K$ and a time varying matrix $M_{t} = ( M^{[1]}_{t}, M^{[2]}_{t}, \dots, M^{[H]}_{t})$. The control action $u_{t}$ at time $t$ by this OC algorithm is given by 
\beq 
u_t = -K x_t + \sum_{k = 1}^H M^{[k]}_{t} w_{t-k}. 
\label{eq:dap}
\eeq 
Thus, the control action is a linear map of the current state and the past disturbances up to a certain history. This property is convenient as it permits efficient optimization of the costs. We note that, since the state is fully observable, the past disturbances can be precisely estimated using the information at time $t$.  

 The parameter $K$ is selected by the OC algorithm as a $(\kappa, \gamma)$-strongly stable linear feedback control matrix for the underlying system. A linear feedback control policy specified by the gain $K$ is $(\kappa, \gamma)$-strongly stable if there exists matrices $L, H$ satisfying $A - BK = HLH^{-1}$ such that the following two conditions are met: $(i)$ $\norm{L} \leq 1 - \gamma$, and $(ii)$  $\norm{K} \leq \kappa, \norm{H}, \norm{H^{-1}} \leq \kappa$. The OC algorithm considers the class $\Pi$ of all $(\kappa, \gamma)$-strongly stable linear feedback controllers for characterizing its regret performance according to \eqref{eq:regret}.

The OC algorithm uses the framework of Online Convex Optimization (OCO) to update the parameters $M_{t}$ at each time step. The key idea of the algorithm is to design a sequence of cost functions $f_{1:T}$ in terms of the parameters $M_{1:T}$ while correctly representing the actual cost incurred by the true cost functions $c_{1:T}$. This is achieved by defining an idealized state $s_{t}$ and idealized control input $a_{t}$ as follows. The idealized state $s_{t}$ is the state the system would have reached if the controller had executed the policy with parameters $(M_{t-H}, \dots, M_{t-1})$ from time step $t-H$ to time step  $t-1$, assuming that the state at $t-H$ is $0$. The idealized action $a_{t}$ is the action that would have been executed at time $t$ if the state observed at time $t$ is $s_t$. We can then  define the idealized cost as $f_t(M_{t-H},\dots,M_{t}) = c_t(s_t,a_t )$. 

The complete OC algorithm proposed in \cite{agarwal2019online} is shown in Algorithm \ref{alg:ocg}. An Online Gradient Descent (OGD) approach updates the parameters $M_{t}$ by the gradient of the idealized cost function. The algorithm requires the specification of a $(\kappa, \gamma)$-strongly stable matrix $K$. Such a matrix can be calculated offline before the task using an Semi-Definite Programming (SDP) relaxation as described in \cite{cohen2018online}. 

\LinesNumberedHidden{\begin{algorithm}[]
\DontPrintSemicolon
\KwInput{Step size $\eta$, parameters $\kappa_B, \kappa, \gamma, T$, $(\kappa,\gamma)$-strongly stable control matrix $K$}

Define $H = {\log{T}}/{\left(\log{\left({1}/{1-\gamma}\right)}\right)}$

Define $\mathcal{M} = \{M = (M^{[1]},\dots,M^{[H]}): \norm{M^{[k]}} \leq \kappa^3 \kappa_B (1-\gamma)^k\}$

Define $g_t(M) = f_t(M,\dots,M)$
 
Initialize $M_1 \in \mathcal{M}$
 
  \For{t = 1,\dots,T}    
   { 
   Choose the action $u_t = -K x_t + \sum_{k = 1}^H M^{[k]}_t w_{t-k}$
   
   Observe the new state $x_{t+1}$, and $w_{t} = x_{t+1} - Ax_t -Bu_t$
   
   Update $M_{t+1} = \tn{Proj}_\mathcal{M}\left( M_t - \eta \nabla g_t(M_t)\right)$
   }

\caption{Online Control (OC) Algorithm}
\label{alg:ocg}
\end{algorithm}}

A regret guarantee of Algorithm \ref{alg:ocg} is provided in \cite{agarwal2019online}:
\begin{theorem}[Theorem 5.1, \cite{agarwal2019online}]
Suppose Assumptions \ref{ass:sys}-\ref{ass:cost} hold, $\eta = \frac{D}{\sqrt{G_f(G_f/2+LH^2)T}}$, and $~ D = \frac{\kappa_B\kappa^3\sqrt{d}}{\gamma}$. Then, under Algorithm \ref{alg:ocg},
\begin{align}
& R_T \leq \frac{3D\sqrt{G_f(G_f/2+LH^2)T}}{2} + \widetilde{\mathcal{O}}(1), ~~ \tn{where} \nonumber \\
& L = 2G\widetilde{D}\kappa_w\kappa_B\kappa^3, ~~ G_f = G\widetilde{D}\kappa_wHd\left(\frac{2\kappa_B\kappa^3}{\gamma} + H \right), \nonumber \\
& \widetilde{D} = \frac{\kappa_w(\kappa^2+H\kappa^2_B\kappa^5)}{\gamma(1-\kappa^2(1-\gamma)^{H+1})} + \frac{\kappa_B\kappa^3\kappa_w}{\gamma}. \nonumber 
\end{align}
\label{thm:ocg}
\end{theorem}

\begin{remark}[Diameter of the domain]
\label{rem:diameter-1}
It can be shown that \cite[Theorem 5.1]{agarwal2019online} the multiplicative constant $D$ in the above regret bound is the diameter of the domain $\mathcal{M}$ of the control policy parameters, i.e., $D =  \max_{M_1, M_2 \in \mathcal{M}} \norm{M_1 - M_2}$. In the next section we show that our meta-learning approach can significantly reduce this multiplicative constant by learning across the tasks. 
\end{remark}

\section{Meta-learning Online Control Algorithm}
\label{sec:metalearner}

\begin{figure}[t!]
\center
\includegraphics[width=0.35\linewidth]{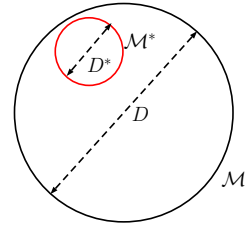}
\caption{Illustrative figure showing the domain $\mathcal{M}$ of the parameters of the online control policies and the  set $\mathcal{M}^{*}$ of the optimal parameters of the control policies corresponding to a set of \textit{similar} tasks. }
\label{fig:tasksimilarity}
\vspace{-0.4cm}
\end{figure}

Our meta-learning online control (M-OC) algorithm builds on the simple, yet a powerful idea of meta-initialization. In the standard OC algorithms, the initialization parameter for the control policy is selected arbitrarily from the domain of possible parameters. So, inevitably the regret guarantee for such algorithms includes a multiplicative constant that is of the order of the radius of the domain (see Remark \ref{rem:diameter-1}), which can be very large in many problems. Similarly, when an independent-learning OC algorithm is applied to a sequence of tasks the parameters of the control policy for each task are initialized arbitrarily ignoring the similarities and the benefit of learning across tasks. When the tasks are similar, the optimal parameters for the individual tasks are closer to each other, and the optimal parameters for the earlier tasks in the sequence can be used to improve the learning in a new upcoming task. Our M-OC algorithm translates this intuitive idea into providing a clever initialization for the control policy for the current task by learning from the previous tasks. This results in a multiplicative constant (in the regret) that is proportional to the diameter  $D^{*}$ of a much smaller subset that contains the parameters of the  optimal control policies of the individual tasks, instead of the diameter of the generic domain. This scenario is illustrated in Fig. \ref{fig:tasksimilarity}, where the diameter $D$ of the original domain $\mathcal{M}$ is significantly larger than $D^{*}$, which is the diameter of the smaller set $\mathcal{M}^{*}$ that contains the optimal parameters corresponding to the similar tasks. Here the diameter $D^{*}$ can be interpreted as the similarity of the sequence of tasks. 

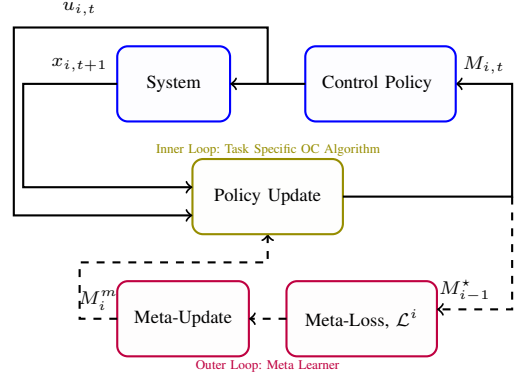
\begin{figure}[t!]
\centering
\begin{tikzpicture}[scale = 0.5]


\draw (0, -4) node[align=center] {\tiny\textcolor{purple}{Outer Loop: Meta Learner}};

\draw [draw=purple, fill=purple, fill opacity = 0.01, rounded corners, thick] (-4, -1.75) rectangle (-0.5, -3.75);
\draw (-2.25, -2.75) node[align=center] {\scriptsize Meta-Update};
\draw [->,thick,dashed] (-4, -2.75) -- (-5,-2.75) -- (-5,-1.25) -- (0,-1.25) -- (0,-0.5);
\draw (-4.5, -2.25) node[align=center] (a) {\scriptsize $M^m_i$};

\draw [draw=purple, fill=purple, fill opacity = 0.01, rounded corners, thick] (4.5, -1.75) rectangle (0.5, -3.75);
\draw (2.5, -2.75) node[align=center] {\scriptsize Meta-Loss, $\mathcal{L}^i$};
\draw [->,thick,dashed] (0.5, -2.75) -- (-0.5,-2.75);

\draw (0, 1.75) node[align=center] {\tiny \textcolor{olive}{Inner Loop: Task Specific OC Algorithm}};


\draw [draw=olive, fill=olive, fill opacity = 0.01, rounded corners, thick] (2, 1.5) rectangle (-2, -0.5);
\draw (0, 0.5) node[align=center] {\scriptsize Policy Update};
\draw [->,thick] (2, 0.5) -- (6.5,0.5) --  (6.5,3.5) --  (5,3.5);
\draw [->,thick,dashed] (6.5, 0.5) -- (6.5,-2.5) -- (4.5,-2.5);
\draw (5.75, 4) node[anchor=center] (a) {\scriptsize $M_{i,t}$};
\draw (5.25, -2) node[anchor=center] (a) {\scriptsize $M^{\star}_{i-1}$};

\draw [draw=blue, fill=blue, fill opacity = 0.01, rounded corners, thick] (-4, 4.5) rectangle (-1, 2.5);
\draw (-2.5, 3.5) node[align=center] {\scriptsize System};
\draw [->,thick] (-4,3.5) -- (-6.5,3.5) -- (-6.5,0.75) -- (-2,0.75); 
\draw (-5, 4) node[anchor=center] (a) {\scriptsize $x_{i, t+1}$};

\draw [draw=blue, fill=blue, fill opacity = 0.01, rounded corners, thick] (5, 4.5) rectangle (1, 2.5);
\draw (3, 3.5) node[align=center] {\scriptsize Control Policy 
};
\draw [->,thick] (1,3.5) -- (-1,3.5 );
\draw [->,thick] (0,3.5) -- (0,5) -- (-6.75,5) -- (-6.75,0) -- (-2,0);
\draw (-5, 5.5) node[anchor=center] (a) {\scriptsize $u_{i, t}$};
 
\end{tikzpicture}
\caption{Meta-Learning Online Control (M-OC) Algorithm Architecture. Solid line: within task signals. Dashed line: signals that are constant within a task but that can change across the tasks.}
\label{fig:metalearner}
\vspace{-0.4cm}
\end{figure}

The architecture of our M-OC algorithm is given in Fig. \ref{fig:metalearner}. The meta-learning in the outer loop provides the meta-initialization for the task specific OC algorithm in the inner loop. The control policy for each specific task is of the same form as the independent learning OC algorithm \eqref{eq:dap}. At the beginning of any task $\tau_{i}$ a $(\kappa, \gamma)$ stabilizing feedback gain matrix $K_i$ for the task $\tau_{i}$ is computed. During the task the algorithm updates the task specific policy parameters $M_{i,t}$ exactly as in Algorithm \ref{alg:ocg}. The control action $u_{i,t}$ is computed using the parameters $M_{i,t}$ and the feedback gain matrix $K_i$ with the same form as the independent learning OC algorithm \eqref{eq:dap}. The difference between the M-OC algorithm and Algorithm \ref{alg:ocg} lies in the initialization of the parameter $M_{i,1}$. In particular, Algorithm \ref{alg:ocg} selects $M_{i,1}$ arbitrarily from the domain $\mathcal{M}$, whereas the outer loop of meta-learner provides the initialization $M^{\mathrm{m}}_{i}$ for each task $\tau_{i}$. 

Specifically, the inner loop updates the control policy parameter $M_{i,t}$ within each task $\tau_{i}$ by
\beq 
M_{i, t+1} = \tn{Proj}_\mathcal{M}\left(M_{i,t} - \nabla g_{i,t}(M_{i,t})\right), ~ M_{i,1} = M^{\mathrm{m}}_{i}.
\label{eq:innerlearner} 
\eeq 
In the outer loop, the meta-learner computes the initialization parameter $M^{\mathrm{m}}_{i}$ for the inner loop as follows. Let $M^{\star}_i$ the optimal parameter in hindsight for task $\tau_{i}$, i.e., 
\beq 
M^{\star}_i = \argmin_{M \in \mathcal{M}} \sum_{t=1}^T g_{i,t}(M).
\label{eq:bestpar}
\eeq 
We note that $M^{\star}_i$ is computable at the end of task $\tau_i$. Given that $g_{i,t}$s are convex functions, finding $M^{\star}_i$ is a convex optimization problem, and thus can be solved efficiently. We define the meta-learner's loss for task $i$ as
\beq 
\mathcal{L}^i(M^{\mathrm{m}}) = \frac{1}{2}\norm{M^{\mathrm{m}} - M^\star_i}^2. 
\label{eq:incloss} 
\eeq 
The meta-learner performs an online gradient descent step to find the initialization  $M^{\mathrm{m}}_{i+1}$ for task $\tau_{i+1}$ as 
\beq 
M^{\mathrm{m}}_{i+1} = \tn{Proj}_\mathcal{M}\left(M^{\mathrm{m}}_i - \frac{1}{i} \nabla \mathcal{L}^i(M^{\mathrm{m}}_i) \right).
\label{eq:outerlearner}
\eeq 

We note that performing the naive initialization $M^{\mathrm{m}}_{i+1} = M^\star_i$ does not improve the regret optimally as this will effectively throw away the information from all the previous tasks. Instead the meta-learner solves an online convex optimization problem with $N$ steps with the cost function at each step $i$ given by $\mathcal{L}^i$. Since the online gradient descent approach solves this problem efficiently with provable guarantees for the regret performance, we adapt this approach as the meta-learning algorithm in the outer loop. We present two variations of the algorithm: (i) a simpler algorithm which assumes knowledge of the diameter $D^{*}$, and (ii) a complete algorithm that does not require the knowledge of $D^{*}$.  

\subsection{Algorithm with the Knowledge of $D^{*}$}

We first present the algorithm with the knowledge of $D^{*}$ for easier understanding of the idea and the technical analysis. The key advantage of this assumption is that we can set the learning rate $\eta$ in the inner loop proportional to $D^{*}$ in addition to updating the meta-initialization according to \eqref{eq:outerlearner}. We emphasize that setting $\eta \propto D^{*}$ is the optimal way to set the rate, which follows from how $\eta$ is set in Theorem \ref{thm:ocg} for the independent learning OC algorithm. The assumption of knowledge of $D^{*}$ simplifies the algorithm, which otherwise requires setting the learning rate adaptively. We present the more general algorithm in the next section. The algorithm with the knowledge of $D^{*}$ is presented below.


\begin{algorithm}[]
\DontPrintSemicolon
\KwInput{Number of tasks $N$, the diameter $D^{*}$, inner loop step size $\eta$, parameters $\kappa_B, \kappa, \gamma, T$}


Define $\mathcal{M} = \{M = (M^{[1]},\dots,M^{[H]}): \norm{M^{[k]}} \leq \kappa^3 \kappa_B (1-\gamma)^k\}$. Initialize $M^{\mathrm{m}}_1 \in \mathcal{M}$ arbitrarily


  \For{i = 1,\dots,N}    
   { 

   For task $\tau_{i}$, set the initialization $M_{i,1} = M^{\mathrm{m}}_i$ for the OC Algorithm (Algorithm \ref{alg:ocg}) in the inner loop 
   
   Execute the OC Algorithm (Algorithm \ref{alg:ocg}) for task $\tau_i$
   
   Compute $M^\star_i$ as in \eqref{eq:bestpar}
   
   Update $M^{\mathrm{m}}_{i+1}$ as in \eqref{eq:incloss}-\eqref{eq:outerlearner} 
   }

\caption{Meta-learning Online Control (M-OC-1) Algorithm }
\label{alg:ml-1}
\end{algorithm}


We now present our main result which characterizes the performance of Algorithm \ref{alg:ml-1}.
\begin{theorem}
Suppose Assumptions \ref{ass:sys}-\ref{ass:cost} hold, and $\eta = \frac{D^{*}}{\sqrt{G_f({G_f}/{2}+LH^2)T}}$. Then, under the M-OC-1 Algorithm (Algorithm \ref{alg:ml-1})
\begin{align}
& {R}^{\mathrm{meta}}_N \leq  \left(\mathcal{O}\left(\frac{\log{N}}{D^{*}N}\right)+\frac{\overline{D}}{2}+D^{*}\right)\sqrt{\widetilde{G}^2T}, \nonumber
\end{align} 
where, $\overline{D}^2 = \frac{1}{N}\sum_{i = 1}^N \left( M^\star_i - \widetilde{M}^\star\right)^2$,  $\widetilde{M}^\star =\frac{1}{N}\sum_{i = 1}^N M^\star_i$, 
 $\widetilde{G}^2 = G_f\left(\frac{G_f}{2}+LH^2\right)$.
\label{thm:metalearner-1}
\end{theorem}

\begin{remark}[Comparison with independent-learning online control algorithm]
\label{rem:compare-with-IL}
Under our M-OC-1 algorithm, when $N$ is sufficiently large, the multiplicative constant in the regret upper bound is approximately equal to $\frac{\overline{D}}{2} + D^{*}$. When the tasks are similar $D^{*} \ll D$, and by definition $\overline{D} \leq D^{*}$. Therefore, when the tasks are similar the regret our algorithm achieves is significantly better compared to the independent learning OC algorithm. This clearly shows that M-OC-1 is able to learn across tasks, which by default the independent learning OC algorithm cannot do. This fact is verified by our numerical simulations also; see Section \ref{sec:meta-simulations}. 
\end{remark}

\begin{remark}[Achievability by meta-learning]
We note that the meta-regret scaling with respect to the duration $T$ of a control task  is $\widetilde{\mathcal{O}}(\sqrt{T})$, which is same as the scaling achieved by the independent learning OC algorithm. This aspect is consistent with the existing theoretical results in online meta-learning \cite{finn2019online, balcan2019provable, khodak2019adaptive}. This is expected, as the meta-learner will never be able to learn an initialization that does not require further adaptation, especially, since the cost functions and the disturbances are arbitrary. Furthermore, as pointed in \cite[Theorem 2.2]{balcan2019provable}, even in the simpler OCO setting, reductions to the multiplicative constant are the best that can be achieved.
\end{remark}

\begin{remark}[Knowledge of $D^{*}$ vs $\mathcal{M}^{*}$]
We emphasize that our algorithm only assumes the knowledge of a scalar $D^{*}$, and not of the entire multi-dimensional set $\mathcal{M}^{*}$. Assuming the knowledge of $\mathcal{M}^{*}$ is not realistic in most practical problems. 
\end{remark}

\subsection{Algorithm without the Knowledge of $D^{*}$}

In this subsection, we present a general version of our algorithm which does not assume the knowledge of $D^{*}$. As mentioned earlier, without the knowledge of $D^{*}$, requires setting the learning rate adaptively.  

Our approach is motivated by the idea proposed in \cite{khodak2019adaptive}, but we  present a simpler algorithm which lends itself to a simpler proof. We set the learning rate for task $\tau_i$ as $\eta = \frac{D_i}{\sqrt{G_f({G_f}/{2}+LH^2)T}}$, where $D_i$ is an estimate of the diameter of the smallest bounding circle of the region $\mathcal{M}^{*}$. We update $D_i$  whenever there is evidence that $D_i$ is smaller that $D^{*}$. The idea is to start $D_i$ from a guess (a small number  $\epsilon$) of $D^{*}$ and increase this guess by a factor $\zeta > 1$ whenever $\norm{M^\star_i - \widetilde{M}^\mathrm{m}_{i-1}} > D_i$, where $\widetilde{M}^\mathrm{m}_i = \frac{1}{i}\sum_{j = 1}^{i} M^\star_i$. The term $\norm{M^\star_i - \widetilde{M}^\mathrm{m}_{i-1}}$ is the deviation of the optimal  parameter for a new task $i$ from the average of the optimal parameters of the previous tasks. Thus, this term is indicative of how smaller $D_i$ is, and thus can be used to increase $D_i$ by comparing with it. In addition, since $\widetilde{M}^\mathrm{m}_{i-1}$ is equal to the output of the meta-learner in Eq. \eqref{eq:outerlearner} with ${M}^\mathrm{m}_1$ set to zero, we use $\widetilde{M}^\mathrm{m}_{i-1}$ itself as the meta-initialization for the task $\tau_i$. The complete algorithm is shown in Algorithm \ref{alg:ml-2}.


\begin{algorithm}[]
\DontPrintSemicolon
\KwInput{Number of tasks $N$, parameters $\kappa_B, \kappa, \gamma, T, \epsilon, \zeta > 1$}


Define $\mathcal{M} = \{M = (M^{[1]},\dots,M^{[H]}): \norm{M^{[k]}} \leq \kappa^3 \kappa_B (1-\gamma)^k\}$. Set $M^{\mathrm{m}}_1$ to the origin. Initialize  $D_1 = \epsilon$, $k = 0$. 

 
  \For{i = 1,\dots,N}    
   { 
   
   Set $ \eta =  \frac{D_i}{\sqrt{G_f({G_f}/{2}+LH^2)T}} $ 

   For task $\tau_{i}$, set the initialization $M_{i,1} = M^{\mathrm{m}}_i$ for the OC Algorithm (Algorithm \ref{alg:ocg}) in the inner loop 
   
   Execute the OC Algorithm (Algorithm \ref{alg:ocg}) for task $\tau_i$
   
   Compute $M^\star_i$ as in \eqref{eq:bestpar}
   
   Set $M^{\mathrm{m}}_{i+1} = \frac{1}{i}\sum_{j = 1}^{i} M^\star_j$
   
   
   \If{$ i > 1$}{
   \If{$\norm{M^\star_i - M^{\mathrm{m}}_{i}} > D_i$}
   {$k \leftarrow k+1$
   }
   }
   $D_{i+1} = \zeta^k\epsilon$}

\caption{Meta-learning Online Control (M-OC-2) Algorithm }
\label{alg:ml-2}
\end{algorithm}

We now present our main result which characterizes the performance of  Algorithm \ref{alg:ml-2}.
\begin{theorem}
Suppose Assumptions \ref{ass:sys}-\ref{ass:cost} hold, $\epsilon < D^{*}$, and $\zeta = (1 + \log(T))/\log(T)$. 
Then, under the M-OC-2 Algorithm (Algorithm \ref{alg:ml-2})
\begin{align}
& {R}^{\mathrm{meta}}_N \leq  \left(\mathcal{O}\left(\frac{\log{N}}{D^{*}N} + \frac{{D}^2}{\epsilon N}\right)+\frac{\overline{D}}{2}+ D^{*} + o_T(1)\right) \sqrt{\widetilde{G}^2T}  \nonumber 
\end{align} 
where, $\overline{D}^2 = \frac{1}{N}\sum_{i = 1}^N \left( M^\star_i - \widetilde{M}^\star\right)^2$,  $\widetilde{M}^\star =\frac{1}{N}\sum_{i = 1}^N M^\star_i$, 
 $\widetilde{G}^2 = G_f\left(\frac{G_f}{2}+LH^2\right)$
\label{thm:metalearner-2}
\end{theorem}

\begin{remark}[Comparison with independent-learning online control algorithm and M-OC-1 algorithm]
\label{rem:compare-with-IL-2}
 Under M-OC-2 algorithm, when $N$ is sufficiently large, the multiplicative constant in the regret upper bound is approximately equal to $\frac{\overline{D}}{2} + D^{*}$. We recall from Remark \ref{rem:compare-with-IL} that $\overline{D} \leq D^{*}$ (by definition), and when the tasks are similar $D^{*} \ll D$. Therefore, when the tasks are similar, we observe that the regret M-OC-2 achieves is significantly better compared to the independent learning OC algorithm. We also observe that the M-OC-2 algorithm has an additional term $\frac{D^{2}}{\epsilon N}$ compared to the M-OC-1 algorithm. This indicates that when the initial guess $\epsilon$ is very small, the number of tasks $N$ that M-OC-2 observes has to be sufficiently large. This is expected as, when $\epsilon$ is much smaller compared to $D^{*}$ meta-learning will necessarily require more experience to improve the initial guess $D_{i} = \epsilon$. 
\end{remark}


\section{Regret Analysis}

In this section, we present a detailed analysis of the M-OC Algorithms \ref{alg:ml-1} and \ref{alg:ml-2}. We first characterize the regret for a single task under these algorithms. The  task regret given by  \eqref{eq:regret} for a task specific policy $\pi^{\mathrm{m}}_{i}$ can be  decomposed  as
\begin{align}
&R^i_T(\pi^{\mathrm{m}}_{i}) \nonumber \\
& = \underbrace{\sum_{t=1}^Tc_{i,t}(x^{\pi^{\mathrm{m}}_{i}}_{i,t},u^{\pi^{\mathrm{m}}_{i}}_{i,t}) - \sum_{t = 1}^T f_{i,t}(M_{i,t-H},\dots,M_{i,t})}_{\tn{Cost Approximation:} ~ R^i_{T,1}} \nonumber \\
& + \underbrace{\sum_{t=1}^T f_{i,t}(M_{i,t-H},. .,M_{i,t}) - \min_{M^{*}} \sum_{t = 1}^T f_{i,t}(M^{*},. .,M^{*})}_{\tn{Policy Regret:} ~ R^i_{T,2}} \nonumber \\
& + \underbrace{\min_{M^{*}} \sum_{t = 1}^T f_{i,t}(M^{*},\dots,M^{*}) - J^{*}_i}_{\tn{Policy Approximation:} ~ R^i_{T,3}},
\label{eq:regretsplit}
\end{align}
where $J^{*}_i = \min_{\pi \in  \Pi} J_i(\pi)$. 

The term $R^i_{T,1}$ is the approximation of the cost by only considering the disturbances upto certain history. The term $R^i_{T,2}$ is the cost difference between the control policy in  \eqref{eq:dap} with $M_{i,t}$ set as the best parameter in hindsight and the optimal policy from the class $\Pi$. The result from \cite[Theorem 5.1]{agarwal2019online} can be used directly to bound the terms $R^i_{T,1}$ and $R^i_{T,3}$.
\begin{lemma}
Under the M-OC Algorithm \ref{alg:ml-1} and Algorithm \ref{alg:ml-2}, the cost approximation term $R^i_{T,1}$ and the policy approximation term $R^i_{T,3}$ are bounded by
\begin{align}
& R^i_{T,1} \leq 2TG\widetilde{D}(1-\gamma)^{H+1}\left(\frac{\kappa_wH\kappa^2_B\kappa^3}{\gamma} + \widetilde{D}\kappa^3\right)\nonumber \\
& = \widetilde{\mathcal{O}}(1), \nonumber \\
& R^i_{T,3} \leq 2TG\widetilde{D}^2\kappa^3(1-\gamma)^{H+1} = \widetilde{\mathcal{O}}(1). \nonumber
\end{align}
\label{lem:r1r3}
\end{lemma}

We note that $H$ is $\mathcal{O}(\log T)$, which results in the final $ \widetilde{\mathcal{O}}(1)$ bound. Also note that $\widetilde{\mathcal{O}}(\cdot)$ hides the poly-log terms. Intuitively the bound for $R^i_{T,1}$ follows from the fact that the idealized cost function as stated earlier is a good approximation of the actual cost. The bound for $R^i_{T,3}$ indicates that the best time invariant control policy of the form \ref{eq:dap} is a good approximation of the best linear feedback policy in hindsight. Next we bound the second term $R^i_{T,2}$. This is the key step in the proof of the regret for the task specific online control, which we then leverage to prove our meta-learning guarantee. This is where our proof differs from the proof of \cite{agarwal2019online}.

\begin{lemma}
Under the M-OC Algorithm \ref{alg:ml-1} and \ref{alg:ml-2}, the policy regret term $R^i_{T,2}$ is bounded by 
\begin{align}
& R^i_{T,2} \leq \frac{\norm{M^{\star}_i-M^{\mathrm{m}}_i}^2}{2\eta} + \frac{TG^2_f\eta}{2} + \eta LH^2G_fT. \nonumber 
\end{align}
\label{lem:r2}
\end{lemma}

\vspace{-0.5cm}
The proof proceeds by splitting $R^i_{T,2}$ to two terms: first term  is the difference between the total idealized cost  and the total cost with the per step cost given by $g_t(M_t)$, and the second term  is the difference between the total cost with the per step cost given by $g_t(M_t)$ and the total idealized cost with $M_{i,t}$ set as the best time invariant parameter in hindsight. The first term is bounded by using the Lipschitz conditions in Assumption \ref{ass:cost} and the second term is bounded by a standard OCO proof methodology. Please see Appendix \ref{sec:pflemr2} for the full proof.

Next, we use the above two lemmas to prove Theorem \ref{thm:metalearner-1} for the M-OC algorithm \ref{alg:ml-1}.

\subsection{Proof of Theorem \ref{thm:metalearner-1}}

By definition
\beq 
{R}^{\mathrm{meta}}_N =  \frac{1}{N} \left( \sum_{i = 1}^N R^i_{T,1} + R^i_{T,2} + R^i_{T,3} \right). \nonumber 
\eeq 
Since, from Lemma \ref{lem:r1r3} $R^i_{T,1} = R^i_{T,2} = \widetilde{\mathcal{O}}(1)$, we neglect these terms and focus only on the remaining term.  
\begin{align}
& {R}^{\mathrm{meta}}_N = \frac{1}{N}\sum_{i = 1}^N R^i_{T,2} + \widetilde{\mathcal{O}}(1)  \nonumber \\
& \stackrel{(a)}{=} \frac{1}{2N\eta}\sum_{i = 1}^N \norm{M^{\star}_i-M^{\mathrm{m}}_i}^2  + \frac{TG^2_f\eta}{2} + \eta LH^2G_fT \nonumber \\
& \stackrel{(b)}{=} \frac{1}{2N\eta}\left(\sum_{i = 1}^N \norm{M^{\star}_i-M^{\mathrm{m}}_i}^2 -  \min_{M^{\mathrm{m}}\in\mathcal{M}}\sum_{i = 1}^N\norm{M^{\star}_i-M^{\mathrm{m}}}^2\right) \nonumber \\
& + \frac{(\Delta^\star)^2}{2\eta} + \frac{TG^2_f\eta}{2} + \eta LH^2G_fT .
\label{eq:pfthm2-eq1}
\end{align}
Here, ($a$) follows from Lemma \ref{lem:r2} and in ($b$) we have used $\Delta^{\star} = \sqrt{\frac{1}{N}\min_{M^{\mathrm{m}} \in \mathcal{M}} \sum_{i = 1}^N \norm{M^{\mathrm{m}} - M^\star_i}^2}$. The key idea now is to bound the term 
\begin{align}
\label{eq:oco-term-to-bound-st1}
\sum_{i = 1}^N \norm{M^{\star}_i-M^{\mathrm{m}}_i}^2 -  \min_{M^{\mathrm{m}}\in\mathcal{M}}\sum_{i = 1}^N\norm{M^{\star}_i-M^{\mathrm{m}}}^2
\end{align}
using the ideas from online convex optimization. For this, consider the OCO problem where the decision at step $i$ is denoted by $M_{i} \in \mathcal{M}$, and the corresponding loss at step $i$ is $\ell_{i}(M_{i})$. The goal of an OCO algorithm is to find a sequence of decisions $M_1, M_2, \dots, M_N$ in order to minimize the regret:
\beq 
\tn{Regret} = R_N = \sum_{i = 1}^{N} \ell_{i}(M_{i}) - \min_{M \in \mathcal{M}}\sum_{i = 1}^{N} \ell_{i}(M).
\eeq 

Consider the case where $\ell_{i}$ is $\alpha_{i}$-strongly convex and $G$-Lipschitz. Then, the following OCO algorithm, which uses the online gradient descent approach, can achieve logarithmic regret \cite[Theorem A.2]{balcan2019provable}:
\beq
M_{i+1} = \tn{Proj}_\mathcal{M}\left(M_{i} - \frac{1}{\sum_{j=1}^{i} \alpha_j} \nabla \ell_{i}(M_{i})\right).
\label{eq:oco-update}
\eeq 

We state this result formally below. 
\begin{lemma}[Theorem A.2, \cite{balcan2019provable}]
{\it Let $\ell_{i} : \mathcal{M} \rightarrow \mathbb{R}$ be a sequence of $\alpha_{i}$-strongly convex and $G$-Lipschitz functions with respect to $\norm{\cdot}$. Then the regret of the online optimization algorithm given in \eqref{eq:oco-update} is $\mathcal{O}(\log(N))$.}
\label{lem:oco-strongconvex}
\end{lemma}

Now, to bound \eqref{eq:oco-term-to-bound-st1}, consider the loss function $\ell_{i}(M^o) = 1/2\norm{M^{\star}_i-M^o}^2$. It is straight forward to show that  $\ell_{i}$ is $1$-strongly convex. It is also Lipschitz inside the set $\mathcal{M}$. Note that the meta-learning step given by \ref{eq:outerlearner} in Algorithm \ref{alg:ml-1}  is indeed the OCO algorithm given in \eqref{eq:oco-update}. Since Eq. \eqref{eq:oco-term-to-bound-st1} represents the regret corresponding to this OCO problem,  Lemma \ref{lem:oco-strongconvex} is applicable here to bound the terms in \eqref{eq:oco-term-to-bound-st1}. Hence, we get
\beq
{R}^{\mathrm{meta}}_N = \frac{\mathcal{O}(\log(N))}{N\eta} + \frac{(\Delta^\star)^2}{2\eta} + \frac{TG^2_f\eta}{2} + \eta LH^2G_fT + \widetilde{\mathcal{O}}(1). 
\label{eq:pfthm2-eq2}
\eeq

The final result follows from substituting the value of $\eta$ and using the fact that $\Delta^\star = \overline{D}$.

\subsection{Proof of Theorem \ref{thm:metalearner-2}}

The steps in the proof are similar to the proof of Theorem \ref{thm:metalearner-1}. By definition
\beq 
{R}^{\mathrm{meta}}_N =  \frac{1}{N} \left( \sum_{i = 1}^N R^i_{T,1} + R^i_{T,2} + R^i_{T,3} \right). \nonumber 
\eeq 
Since, the learning rate is set differently in each task $\tau_i$, we denote the learning rate in task $\tau_i$ by $\eta_i$. Since $R^i_{T,1} = R^i_{T,2} = \widetilde{\mathcal{O}}(1)$ from Lemma \ref{lem:r1r3}, we focus only on the remaining term. 
\begin{align}
& {R}^{\mathrm{meta}}_N = \frac{1}{N}\sum_{i = 1}^N R^i_{T,2}  \nonumber \\
& = \frac{1}{N}\sum_{i = 1}^N \left( \frac{\norm{M^{\star}_i-M^{\mathrm{m}}_i}^2}{2\eta_i}  + \frac{TG^2_f\eta_i}{2} + \eta_i LH^2G_fT\right) \nonumber 
\end{align}

Here the last equality follows from Lemma \ref{lem:r2}. The following observations hold: (i) the average $\widetilde{M}^{\mathrm{m}}_i = \frac{1}{i}\sum_{j=1}^{i}M^\star_j $ is a convex combination and thus lies within the smallest bounding circle of $\mathcal{M}^{*}$. Thus, given the fact that $\widetilde{M}^{\mathrm{m}}_{i-1} = M^{\mathrm{m}}_i$ for $i > 1$, $M^{\mathrm{m}}_{i}$ is always be within $D^{*}$ distance from $M^\star_i$ for all $i$s. 

Given how $D_i$ is increased from one task to the next, it follows from the previous observation that there are at the most $\floor{\log_\zeta(\frac{D^{*}}{\epsilon})}$ tasks after $i = 1$ when $\norm{M^\star_i - M^{\mathrm{m}}_{i}} > D_i$. We index such instances by $k$ and denote the corresponding task indices by $i_k$.

Lets define an alternate sequence in which $\widetilde{D}_1 = D^{*}$, $\widetilde{D}_i = D_i$ when $\norm{M^\star_i - M^{\mathrm{m}}_{i}} \leq D_i$ for any $i > 1$, and $\widetilde{D}_i = D^{*}$ otherwise. Let $\widetilde{G}^2 = \left(\frac{G^2_f}{2} + LH^2G_f\right)$. Then it follows that $\eta_i = \frac{D_i}{\widetilde{G}\sqrt{T}}$. Then 
\begin{align}
& {R}^{\mathrm{meta}}_N  \leq \frac{1}{N}\left( \frac{\norm{M^{\star}_1-M^{\mathrm{m}}_{1}}^2 }{2D_1} + D_1 \right) \sqrt{\widetilde{G}^2T}\nonumber \\ 
& + \frac{1}{N}\sum_{i = 2}^N \left( \frac{\norm{M^{\star}_i-M^{\mathrm{m}}_{i}}^2}{2D_i} + D_i\right) \sqrt{\widetilde{G}^2T} \nonumber \\
& \stackrel{(a)}{\leq} \frac{D^2}{2N\epsilon}\sqrt{\widetilde{G}^2T} \nonumber \\
& + \frac{1}{N}\sum_{i = 1}^N \left( \frac{\norm{M^{\star}_i-M^{\mathrm{m}}_{i}}^2}{2\widetilde{D}_i} + \widetilde{D}_i\right) \sqrt{\widetilde{G}^2T}  \nonumber \\
& + \frac{1}{N}\sum_{k = 0}^{\floor{\log_\zeta(\frac{D^{*}}{\epsilon})}} \left( \frac{\norm{M^{\star}_{i_k}-M^{\mathrm{m}}_{i_k}}^2}{2\zeta^k\epsilon}  + \zeta^k\epsilon\right) \sqrt{\widetilde{G}^2T}. \nonumber
\end{align}

Here $(a)$ follows from adding additional terms for all those tasks when $\widetilde{D}_i \neq D_i$, which by definition occurs when $i = 1$ and when $\norm{M^\star_{i} - M^{\mathrm{m}}_{i}} > D_{i}$ for $i > 1$. 

We make some observations. By definition, $\widetilde{D}_i \geq \norm{M^{\star}_i-M^{\mathrm{m}}_{i}}$ when $i > 1$. Consider the function $g(x) = \frac{B^2}{x} + x$. We observe that this function is increasing for $x \geq B$. We also observe that $D_i \leq \zeta D^{*}$. With these observations we can simplify the bound to ${R}^{\mathrm{meta}}_N$ as
\begin{align}
& {R}^{\mathrm{meta}}_N \leq \frac{D^2}{2N\epsilon}\sqrt{\widetilde{G}^2T} \nonumber \\
& + \frac{1}{N}\sum_{i = 1}^N \left( \frac{\norm{M^{\star}_i-M^{\mathrm{m}}_{i}}^2}{2D^{*}} + \zeta D^{*}\right) \sqrt{\widetilde{G}^2T}  \nonumber \\
& + \frac{1}{N}\sum_{k = 0}^{\floor{\log_\zeta(\frac{D^{*}}{\epsilon})}} \left( \frac{\norm{M^{\star}_{i_k}-M^{\mathrm{m}}_{i_k}}^2}{2\zeta^k\epsilon}  + \zeta^k\epsilon\right) \sqrt{\widetilde{G}^2T}. \nonumber
\end{align}

Next we bound the last term. We note that by definition $\norm{M^{\star}_{i_k}-M^{\mathrm{m}}_{i_k}} \leq D^{*}$ for all $i > 1$ and by definition $i_k > 1$. Let $K = \floor{\log_\zeta(\frac{D^{*}}{\epsilon})}$. Therefore,
\begin{align}
& \frac{\sqrt{\widetilde{G}^2T}}{N}\sum_{k = 0}^{K} \left( \frac{\norm{M^{\star}_{i_k}-M^{\mathrm{m}}_{i_k}}^2}{2\zeta^k\epsilon}  + \zeta^k\epsilon\right) \nonumber \\
& \leq \frac{\sqrt{\widetilde{G}^2T}}{N}\sum_{k = 0}^{K} \left( \frac{{D^{*}}^2}{2\zeta^k\epsilon} + \zeta^k\epsilon\right) \nonumber \\
& = \frac{\sqrt{\widetilde{G}^2T}}{N}\left( \frac{{D^{*}}^2(\zeta^{K+1}-1)}{2\zeta^{K}(\zeta - 1)\epsilon} + \frac{\epsilon(\zeta^{K+1}-1)}{\zeta - 1}\right) \nonumber \\
& = \mathcal{O}\left( \frac{{D^{*}}^2}{\epsilon N}\right)\sqrt{\widetilde{G}^2T}. \nonumber
\end{align}

Next we bound the second term. Let $\widetilde{M}^\star := \frac{1}{N}\sum_{i=1}^N M^\star_i$. Adding and subtracting $\norm{M^\star_i - \widetilde{M}^\star}$ for each $i$, we get
\begin{align}
&  \frac{1}{N}\sum_{i = 1}^N \left( \frac{\norm{M^{\star}_i-{M}^{\mathrm{m}}_{i}}^2}{2D^{*}} + \zeta D^{*}\right) \sqrt{\widetilde{G}^2T}  \nonumber \\
& = \frac{\sqrt{\widetilde{G}^2T}}{2D^{*}N}\sum_{i = 1}^N \left( \norm{M^{\star}_i-{M}^{\mathrm{m}}_{i}}^2-\norm{M^{\star}_i-\widetilde{M}^{\star}}^2\right)  \nonumber \\
& + \frac{\sqrt{\widetilde{G}^2T}}{N}\sum_{i = 1}^N \left(\frac{\norm{M^{\star}_i-\widetilde{M}^{\star}}^2}{2D^{*}} + \zeta D^{*}\right)  \nonumber \\
& \stackrel{(d)}{\leq} \left( \frac{\mathcal{O}(\log(N))}{N} + \frac{\overline{D}}{2} + \zeta D^{*}\right)\sqrt{\widetilde{G}^2T}.
\end{align}

Here $(d)$ follows from (i) \[\widetilde{M}^{\star} = \arg\min_{M \in \mathcal{M}} \frac{1}{N}\sum_{i=1}^N \norm{M^\star_i - M}^2, \nonumber \] (ii) $\sum_{i = 1}^N \left( \norm{M^{\star}_i-{M}^{\mathrm{m}}_{i}}^2-\norm{M^{\star}_i-\widetilde{M}^{\star}}^2\right)$ is the regret for meta-learning given (i) and Lemma \ref{lem:oco-strongconvex}, and (iii) by definition of $\overline{D}$. The final result follows from combining all terms.

\section{Numerical Experiments}
\label{sec:meta-simulations}

In this section, we present numerical experiments to demonstrate the benefits of our proposed meta-learning online control algorithm. We consider only the M-OC-1 algorithm for the simplicity of illustration. In our experiments, each task $\tau_{i}$ is the problem of regulating a linear dynamical system given in \ref{eq:syseq} with dimensions $n = 2, m = 1$. The system model $A_{i}$ in each task $\tau_{i}$ is selected as a random matrix: a perturbation around a nominal matrix. In particular, we set $A_{i} = \frac{1}{2n} I + \frac{1}{5n} W_{i}$, where $W_{i}$ is a random matrix with the value of each element generated uniformly from the interval $[0, 1]$. This structure implicitly incorporates the idea of task similarity. The cost functions $c_{i,t}$s are selected as quadratic cost functions $c_{i,t}(x,u) = x^\top Q_tx + u^\top R_t u$, where $Q_t$ and $R_t$ are randomly chosen diagonal matrices with each diagonal element chosen randomly from the range $[0.375, 0.625]$. The other parameters are selected as $\kappa_a = \kappa_b = \kappa_w = 1, \kappa = \sqrt{nm}, \gamma = 0.5$. 

In our experiments, we compare the performance of our M-OC algorithm with the following benchmarks: \\
$(i)$ \textit{Non-adaptive control algorithm} which employs the control policy $u_{i,t} = - K_{i} x_{t}$, where $K_{i}$ is a stabilizing controller for task $\tau_{i}$ with system parameter $\theta_{i} = [A_{i}, B_{i}]$. We select $K_{i}$ by solving a standard linear matrix inequality (LMI) for finding a stabilizing controller. We call this non-adaptive control because the control policy is invariant over the duration of the control tasks. Moreover, there is no learning across the tasks.  \\
$(ii)$ \textit{Independent-learning online control algorithm} employs the task specific OC algorithm (Algorithm \ref{alg:ocg}) independently to each control task. While this approach is capable of learning within a task, it does not perform any meta-learning across the tasks. 

Different from these benchmarks, our M-OC algorithm can learn within and across the tasks.

\begin{figure}
\center
\includegraphics[width=0.8\linewidth]{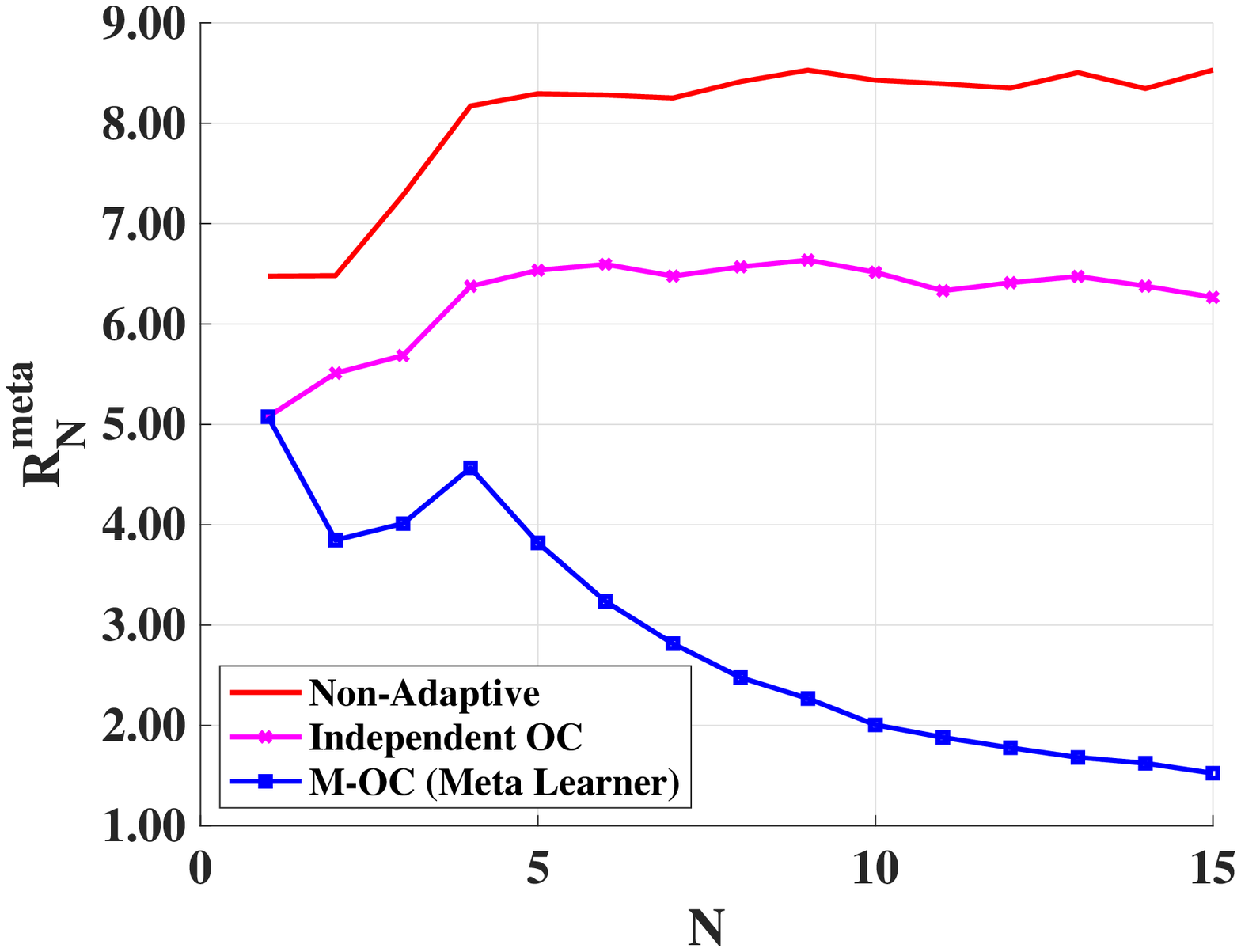}
\caption{Plot of $R^{\mathrm{meta}}_N$ versus the number for tasks $N$.}
\label{fig:costvstask}
\end{figure}

\begin{figure}
\center
\includegraphics[width=0.8\linewidth]{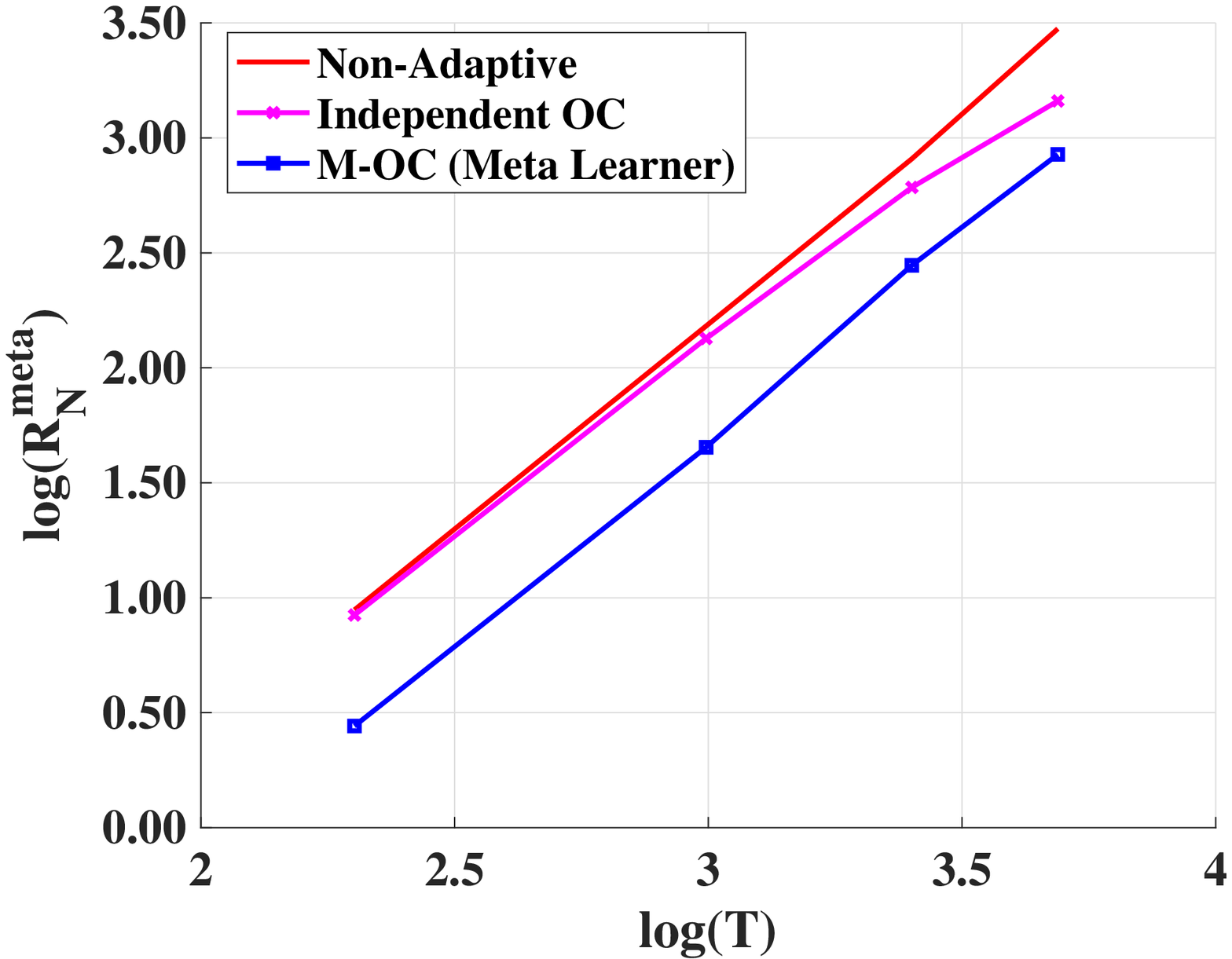}
\caption{Plot of $\log{R^{\mathrm{meta}}_N}$ vs $\log{T}$.}
\label{fig:avgregvsT}
\end{figure}

Figure \ref{fig:costvstask} shows the meta-regret ${R}^{\mathrm{meta}}_N$ as a function of the number of tasks $N$ with $T = 25$ for all tasks. Note that meta-regret is equivalent to the average (averaged over the tasks) cumulative regret of the tasks; see \eqref{eq:averageregret}. Since the non-adaptive control algorithm and the independent-learning OC algorithm do not perform any learning across the tasks, their meta-regret does not improve with the number of tasks. In stark difference, the meta-regret of our M-OC algorithm decreases with the number of tasks; see Remark \ref{rem:compare-with-IL} also. This is because our M-OC algorithm is designed to perform meta-learning across the tasks. This clearly demonstrates the superior performance of the M-OC algorithm over the benchmarks without meta-learning.  

Figure \ref{fig:avgregvsT} shows the variation of the meta-regret with  $N=15$ tasks as a function of the duration $T$ of each control task. We see that, when the task duration is small, the M-OC outperforms independent learning OC by a notable margin. This indeed is the very purpose meta-learning, i.e., to improve adaptation when the data or experience available for online learning is limited.

\section{Conclusion}
\label{sec:conc}

In this paper, we address the problem of developing a meta-learning online control algorithm for a sequence of similar control tasks. We focus on the setting where each task is the problem of controlling a linear dynamical system with arbitrary disturbances and arbitrarily time varying cost functions. We propose a meta-learning online control algorithm that provably achieves a superior performance compared to the standard online control algorithm which does not use meta-learning. We also present numerical experiments to demonstrate the superior performance of our algorithm.  In the future work, we plan to extend this approach to the setting where the system parameters $\theta_{i}$s are also unknown.

\bibliographystyle{IEEEtran} 
\bibliography{Refs.bib}

\begin{thebibliography}{10}
\providecommand{\url}[1]{#1}
\csname url@samestyle\endcsname
\providecommand{\newblock}{\relax}
\providecommand{\bibinfo}[2]{#2}
\providecommand{\BIBentrySTDinterwordspacing}{\spaceskip=0pt\relax}
\providecommand{\BIBentryALTinterwordstretchfactor}{4}
\providecommand{\BIBentryALTinterwordspacing}{\spaceskip=\fontdimen2\font plus
\BIBentryALTinterwordstretchfactor\fontdimen3\font minus
  \fontdimen4\font\relax}
\providecommand{\BIBforeignlanguage}[2]{{%
\expandafter\ifx\csname l@#1\endcsname\relax
\typeout{** WARNING: IEEEtran.bst: No hyphenation pattern has been}%
\typeout{** loaded for the language `#1'. Using the pattern for}%
\typeout{** the default language instead.}%
\else
\language=\csname l@#1\endcsname
\fi
#2}}
\providecommand{\BIBdecl}{\relax}
\BIBdecl

\bibitem{thrun2012learning}
S.~Thrun and L.~Pratt, \emph{Learning to learn}.\hskip 1em plus 0.5em minus
  0.4em\relax Springer Science \& Business Media, 2012.

\bibitem{finn2017model}
C.~Finn, P.~Abbeel, and S.~Levine, ``Model-agnostic meta-learning for fast
  adaptation of deep networks,'' in \emph{International Conference on Machine
  Learning (ICML)}, 2017.

\bibitem{ravi2017optimization}
S.~Ravi and H.~Larochelle, ``Optimization as a model for few-shot learning,''
  in \emph{International Conference on Learning Representations ({ICLR})},
  2017.

\bibitem{nichol2018first}
A.~Nichol, J.~Achiam, and J.~Schulman, ``On first-order meta-learning
  algorithms,'' \emph{arXiv preprint arXiv:1803.02999}, 2018.

\bibitem{wang2020frustratingly}
X.~Wang, T.~Huang, J.~Gonzalez, T.~Darrell, and F.~Yu, ``Frustratingly simple
  few-shot object detection,'' in \emph{International Conference on Machine
  Learning}, 2020.

\bibitem{bragg2021flex}
J.~Bragg, A.~Cohan, K.~Lo, and I.~Beltagy, ``Flex: Unifying evaluation for
  few-shot nlp,'' in \emph{Neural Information Processing Systems (NeurIPS)},
  2021.

\bibitem{NagabandiCLFALF19}
A.~Nagabandi, I.~Clavera, S.~Liu, R.~S. Fearing, P.~Abbeel, S.~Levine, and
  C.~Finn, ``Learning to adapt in dynamic, real-world environments through
  meta-reinforcement learning,'' in \emph{International Conference on Learning
  Representations (ICLR)}, 2019.

\bibitem{li2010contextual}
L.~Li, W.~Chu, J.~Langford, and R.~E. Schapire, ``A contextual-bandit approach
  to personalized news article recommendation,'' in \emph{International
  conference on World Wide Web}, 2010, pp. 661--670.

\bibitem{lesort2020continual}
T.~Lesort, V.~Lomonaco, A.~Stoian, D.~Maltoni, D.~Filliat, and
  N.~D{\'\i}az-Rodr{\'\i}guez, ``Continual learning for robotics: Definition,
  framework, learning strategies, opportunities and challenges,''
  \emph{Information fusion}, vol.~58, pp. 52--68, 2020.

\bibitem{petrivc2014online}
T.~Petri{\v{c}}, A.~Gams, L.~{\v{Z}}lajpah, and A.~Ude, ``Online learning of
  task-specific dynamics for periodic tasks,'' in \emph{2014 IEEE/RSJ
  International Conference on Intelligent Robots and Systems}.\hskip 1em plus
  0.5em minus 0.4em\relax IEEE, 2014, pp. 1790--1795.

\bibitem{alambeigi2018robust}
F.~Alambeigi, Z.~Wang, R.~Hegeman, Y.-H. Liu, and M.~Armand, ``A robust
  data-driven approach for online learning and manipulation of unmodeled 3-d
  heterogeneous compliant objects,'' \emph{IEEE Robotics and Automation
  Letters}, vol.~3, no.~4, pp. 4140--4147, 2018.

\bibitem{romeres2019derivative}
D.~Romeres, M.~Zorzi, R.~Camoriano, S.~Traversaro, and A.~Chiuso,
  ``Derivative-free online learning of inverse dynamics models,'' \emph{IEEE
  Transactions on Control Systems Technology}, vol.~28, no.~3, pp. 816--830,
  2019.

\bibitem{tesfazgi2021personalized}
S.~Tesfazgi, A.~Lederer, J.~F. Kunz, A.~J. Ord{\'o}{\~n}ez-Conejo, and
  S.~Hirche, ``Personalized rehabilitation robotics based on online learning
  control,'' \emph{arXiv preprint arXiv:2110.00481}, 2021.

\bibitem{kalathil2015online}
D.~Kalathil and R.~Rajagopal, ``Online learning for demand response,'' in
  \emph{Annual Allerton Conference on Communication, Control, and Computing
  (Allerton)}, 2015, pp. 218--222.

\bibitem{lin2012online}
M.~Lin, Z.~Liu, A.~Wierman, and L.~L. Andrew, ``Online algorithms for
  geographical load balancing,'' in \emph{International Green Computing
  Conference (IGCC)}, 2012.

\bibitem{shalev2011online}
S.~Shalev-Shwartz \emph{et~al.}, ``Online learning and online convex
  optimization,'' \emph{Foundations and trends in Machine Learning}, vol.~4,
  no.~2, pp. 107--194, 2011.

\bibitem{hazan2019introduction}
E.~Hazan, ``Introduction to online convex optimization,'' \emph{arXiv preprint
  arXiv:1909.05207}, 2019.

\bibitem{dean2018regret}
S.~Dean, H.~Mania, N.~Matni, B.~Recht, and S.~Tu, ``Regret bounds for robust
  adaptive control of the linear quadratic regulator,'' in \emph{Neural
  Information Processing Systems (NeurIPS)}, 2018.

\bibitem{mania2019certainty}
H.~Mania, S.~Tu, and B.~Recht, ``Certainty equivalence is efficient for linear
  quadratic control,'' in \emph{Neural Information Processing Systems
  (NeurIPS)}, 2019.

\bibitem{agarwal2019online}
N.~Agarwal, B.~Bullins, E.~Hazan, S.~Kakade, and K.~Singh, ``Online control
  with adversarial disturbances,'' \emph{International Conference on Machine
  Learning (ICML)}, pp. 111--119, 2019.

\bibitem{simchowitz2020improper}
M.~Simchowitz, K.~Singh, and E.~Hazan, ``Improper learning for non-stochastic
  control,'' \emph{Conference on Learning Theory (COLT)}, pp. 3320--3436, 2020.

\bibitem{finn2019online}
C.~Finn, A.~Rajeswaran, S.~Kakade, and S.~Levine, ``Online meta-learning,'' in
  \emph{International Conference on Machine Learning (ICML)}, 2019, pp.
  1920--1930.

\bibitem{balcan2019provable}
M.-F. Balcan, M.~Khodak, and A.~Talwalkar, ``Provable guarantees for
  gradient-based meta-learning,'' \emph{International Conference on Machine
  Learning (ICML)}, pp. 424--433, 2019.

\bibitem{khodak2019adaptive}
M.~Khodak, M.-F.~F. Balcan, and A.~S. Talwalkar, ``Adaptive gradient-based
  meta-learning methods,'' \emph{Neural Information Processing Systems
  (NeurIPS)}, 2019.

\bibitem{cohen2019learning}
A.~Cohen, T.~Koren, and Y.~Mansour, ``Learning linear-quadratic regulators
  efficiently with only $\sqrt{T}$ regret,'' \emph{International Conference on
  Machine Learning (ICML)}, pp. 1300--1309, 2019.

\bibitem{simchowitz2020naive}
M.~Simchowitz and D.~Foster, ``Naive exploration is optimal for online lqr,''
  \emph{International Conference on Machine Learning (ICML)}, pp. 8937--8948,
  2020.

\bibitem{hazan2020nonstochastic}
E.~Hazan, S.~Kakade, and K.~Singh, ``The nonstochastic control problem,''
  \emph{Algorithmic Learning Theory}, pp. 408--421, 2020.

\bibitem{sastry2011adaptive}
S.~Sastry and M.~Bodson, \emph{Adaptive control: stability, convergence and
  robustness}.\hskip 1em plus 0.5em minus 0.4em\relax Dover Publications, 2011.

\bibitem{aastrom2013adaptive}
K.~J. {\AA}str{\"o}m and B.~Wittenmark, \emph{Adaptive control}.\hskip 1em plus
  0.5em minus 0.4em\relax Courier Corporation, 2013.

\bibitem{ioannou2012robust}
P.~A. Ioannou and J.~Sun, \emph{Robust adaptive control}.\hskip 1em plus 0.5em
  minus 0.4em\relax Dover Publications, 2012.

\bibitem{ZhouDoyleGlover-RobustControl-Book}
K.~Zhou, J.~Doyle, and K.~Glover, \emph{Robust and optimal control}.\hskip 1em
  plus 0.5em minus 0.4em\relax Prentice hall, 1996.

\bibitem{cohen2018online}
A.~Cohen, A.~Hasidim, T.~Koren, N.~Lazic, Y.~Mansour, and K.~Talwar, ``Online
  linear quadratic control,'' in \emph{International Conference on Machine
  Learning (ICML)}, 2018, pp. 1029--1038.

\end{thebibliography}

\begin{appendices}

\section{Proof of Lemma \ref{lem:r2} }
\label{sec:pflemr2}
In the following, for convenience we drop the subscript $i$. We first introduce \cite[Lemma 5.6]{agarwal2019online} and \cite[Lemma 5.7]{agarwal2019online} which are useful in our proof. 

\begin{lemma}[Lemma 5.6, \cite{agarwal2019online}]
Consider two policy sequences $(M_{t-H}, \dots, M_{t-k}, \dots, M_t)$ and $(M_{t-H}, \dots, \tilde{M}_{t-k}, \dots, M_t)$ which differ only in the policy at time $t-k$, where $k \in \{0,1,\dots,H\}$. Then,
\begin{align}
& \vert f_t(M_{t-H} \dots M_{t-k} \dots M_t) -  f_t(M_{t-H} \dots \tilde{M}_{t-k} \dots M_t)\vert \nonumber \\
& \leq L \norm{M_{t-k} - \tilde{M}_{t-k}}. \nonumber
\end{align}
\label{lem:lipschitzbound}
\end{lemma}

\begin{lemma}[Lemma 5.7, \cite{agarwal2019online}]
For all $M$ such that $\norm{M^{[j]}} \leq \kappa_B\kappa^3(1-\gamma)^j, ~ \forall ~ j \in \{1,\dots,H\}$, we have that
\beq
\norm{\nabla_M f_t(M,\dots,M)} \leq G_f. \nonumber
\eeq
\label{lem:gradbound}
\end{lemma}

We now give the main proof. We can split the policy regret term $R_{T,2}$ as
\begin{align}
& R_{T,2} \nonumber \\
& = \sum_{t=1}^T f_t(M_{t-H},\dots,M_t) - \min_{M^{*}} \sum_{t = 1}^T f_t(M^{*},\dots,M^{*}) \nonumber \\
& = \underbrace{\sum_{t=1}^T f_t(M_{t-H},\dots,M_t) - \sum_{t=1}^T f_t(M_{t},\dots,M_t)}_{\tn{Term I}} \nonumber \\
& +  \underbrace{\sum_{t=1}^T f_t(M_{t},\dots,M_t) - \min_{M^{*}} \sum_{t = 1}^T f_t(M^{*},\dots,M^{*})}_{\tn{Term II}}. \nonumber
\end{align}

First we bound Term I. 
\begin{align}
& \tn{Term I} \stackrel{(a)}{\leq} L \sum_{t=1}^T \sum_{j = 1}^H \norm{M_t - M_{t-j}} \nonumber\\
& \stackrel{(b)}{\leq} L \sum_{t=1}^T \sum_{j = 1}^H \sum_{l = 1}^j \norm{M_{t-l+1} - M_{t-l}} \nonumber \\
& \stackrel{(c)}{\leq} L\eta \sum_{t=1}^T \sum_{j = 1}^H \sum_{l = 1}^j \norm{\nabla f_{t-l}(M_{t-l})} \stackrel{(d)}{\leq} TLH^2\eta G_f. \nonumber 
\end{align}

Here ($a$) follows from subtracting and adding $f_t(M_{t-H},\dots,M_{t-j},M_t,\dots,M_t)$ for all $j \in \{2,\dots,H\}$ and for all $t$, applying triangle inequality, and Lemma \ref{lem:lipschitzbound}, ($b$) follows from adding and subtracting $M_{t-l}$, for all $l \in \{1,\dots,j-1\}$, inside the norm for all $j$ and $t$, and applying triangle inequality, ($c$) follows from Eq. \eqref{eq:outerlearner} and ($d$) follows from applying Lemma \ref{lem:gradbound} and summing all terms. 

Next we bound Term II. Since $c_t$ is convex and $s_t$ and $a_t$ are linear in $M_{t-j}$ for all $j \in \{0,\dots,H\}$, it follows that $f_t(M,\dots,M)$ is convex in $M$. In the following, we use the notation $f_t(M,\dots,M) = g_t(M)$. In the steps to follow, we use vectorial expansion for the matrices and the gradients to simplify the algebraic manipulation. We denote the $\nabla^v g_t(M_t)$ as the vectorial expansion of the gradient of $g_t(M_t)$ and $M^v_t$ and $M^{\star,v}$ as the vectorial expansion of the matrices $M_t$ and $M^{\star}$. Since $g_t(M)$ is convex in $M$, we get that 
\begin{align}
& \tn{Term II} = \sum_{t=1}^T g_t(M_t) - \min_{M^{*} \in \mathcal{M}} \sum_{t=1}^T g_t(M^{*}) \nonumber \\
& \leq \sum_{t=1}^T \nabla^v g_t(M_t)^{\top} (M^v_t - M^{\star,v}_i). 
\label{eq:pflem2-eq1}
\end{align}

Now
\begin{align}
 & \norm{M^v_t - M^{\star,v}}^2 - \norm{M^v_{t+1} - M^{\star,v}}^2 \stackrel{(e)}{=} \norm{M^v_t - M^{\star,v}}^2 \nonumber \\
 & - \norm{\tn{Proj}\left(M^v_t - \eta \nabla^v g_t(M_t)\right) - M^{\star,v}}^2 \nonumber \\
 & \stackrel{(f)}{\geq} \norm{M^v_t - M^{\star,v}}^2 
 - \norm{M^v_t - \eta \nabla^v g_t(M_t) - M^{\star,v}}^2 \nonumber \\
 & \stackrel{(g)}{\geq} 2\eta \nabla^v g_t(M_t)^{\top} (M^v_t - M^{\star,v})  -\eta^2 \norm{\nabla^v g_t(M_t)}^2.
 \label{eq:pflem2-eq2}
\end{align}

Here ($e$) follows using the meta-update rule Eq. \eqref{eq:innerlearner}, ($f$) follows from the trivial fact that projection to a set decreases the euclidean distance to any element within the set, ($g$) follows from just expanding the second term and canceling out the identical terms. 

Then from Eq. \eqref{eq:pflem2-eq2} it follows that
\begin{align}
& \nabla^v g_t(M_t)^{\top} (M^v_t - M^{\star,v}) \nonumber \\
& \leq \frac{1}{2\eta }\left(\norm{M^v_t - M^{\star,v}}^2 - \norm{M^v_{t+1} - M^{\star,v}}^2 \right) +\frac{\eta G_f}{2}.\nonumber 
\end{align}

Then combining Eq. \eqref{eq:pflem2-eq1} and the previous equation, and summing over $t$ we get that
\begin{align}
& \tn{Term II} \leq \frac{1}{2\eta}\left(\norm{M^v_1 - M^{\star,v}}^2 - \norm{M^v_{T+1} - M^{\star,v}}^2 \right) \nonumber \\
& + \frac{T\eta G_f}{2} \leq \frac{1}{2\eta}\norm{M^v_1 - M^{\star,v}}^2 + \frac{T\eta G_f}{2} \nonumber \\
& \stackrel{(h)}{=} \frac{1}{2\eta}\norm{M_1 - M^{\star}}^2 + \frac{T\eta G_f}{2} \nonumber \\
& \stackrel{(i)}{=} \frac{1}{2\eta}\norm{M^{\mathrm{m}} - M^{\star}}^2 + \frac{T\eta G_f}{2}.
\end{align}

Here ($h$) follows from the fact that square of the Frobenious norm of a matrix is the square of the Euclidean norm of its vectorial expansion, ($i$) follows from the fact that $M_1$ in task $\tau_i$ is equal to $M^{\mathrm{m}}_i$. Combining the bounds for Term I and Term II we get the final result.

\end{appendices} 

\begin{IEEEbiography}
{Deepan Muthirayan}
is currently a Post-doctoral Researcher in the department of Electrical Engineering and Computer Science at the University of California at Irvine. He obtained his Phd from the University of California at Berkeley (2016) and B.Tech/M.tech degree from the Indian Institute of Technology Madras (2010). His doctoral thesis work focused on market mechanisms for integrating demand flexibility in energy systems. Before his term at UC Irvine he was a post-doctoral associate at Cornell University where his work focused on online scheduling algorithms for managing demand flexibility. His current research interests include control theory, machine learning, learning for control, online learning, game theory, and their application to smart systems.
\end{IEEEbiography}

\begin{IEEEbiography}
{Dileep Kalathil} (Senior Member, IEEE) received his Ph.D. degree from the University of Southern California (USC) in 2014. From 2014 to 2017, he was a Postdoctoral Researcher with the Department of Electrical Engineering and Computer Sciences, University of California at Berkeley. He is currently an Assistant Professor with the Department of Electrical and Computer Engineering, Texas A\& M University. His main research focus is on reinforcement learning theory and algorithms, with applications in energy systems, communication networks and mobile robotics. He was a recipient of the Best Academic Performance from the EE Department, IIT Madras and the Best Ph.D. Dissertation Prize in the USC Department of Electrical Engineering, NSF CRII Award in 2019 and NSF CAREER award in 2021. 
\end{IEEEbiography}

\begin{IEEEbiography}
{Pramod Khargonekar} received B. Tech. Degree in electrical engineering in 1977 from the Indian Institute of Technology, Bombay, India, and M.S. degree in mathematics in 1980 and Ph.D. degree in electrical engineering in 1981 from the University of Florida, respectively. He was Chairman of the Department of Electrical Engineering and Computer Science from 1997 to 2001 and also held the position of Claude E. Shannon Professor of Engineering Science at The University of Michigan.  From 2001 to 2009, he was Dean of the College of Engineering and Eckis Professor of Electrical and Computer Engineering at the University of Florida till 2016. After serving briefly as Deputy Director of Technology at ARPA-E in 2012-13, he was appointed by the National Science Foundation (NSF) to serve as Assistant Director for the Directorate of Engineering (ENG) in March 2013, a position he held till June 2016. Currently, he is Vice Chancellor for Research and Distinguished Professor of Electrical Engineering and Computer Science at the University of California, Irvine. His research and teaching interests are centered on theory and applications of systems and control. He has received numerous honors and awards including IEEE Control Systems Award, IEEE Baker Prize, IEEE CSS Axelby Award, NSF Presidential Young Investigator Award, AACC Eckman Award, and is a Fellow of IEEE, IFAC, and AAAS.

\end{IEEEbiography}
\end{document}